\documentclass{article}
\usepackage{spconf,amsmath,graphicx}

\usepackage{enumitem}
\setlist{nosep, leftmargin=14pt}

\usepackage{mwe} 

\usepackage{booktabs}
\usepackage{amssymb,amsfonts}
\usepackage{algorithmic}
\usepackage{amsthm}
\usepackage{multirow}

\title{MLIP: Medical Language-Image Pre-training with\protect\\Masked Local Representation Learning}
\name{
\begin{tabular}{c}
    {Jiarun Liu$^{1,2,3}$ \qquad Hong-Yu Zhou$^{4}$ \qquad Cheng Li$^{1}$} \\
    {Weijian Huang$^{1,2,3}$ \qquad Hao Yang$^{1,2,3}$ \qquad Yong Liang$^{2,5}$ \qquad Shanshan Wang$^{1,2,\dagger}$\thanks{$^{\dagger}$ Corresponding author. ss.wang@siat.ac.cn}}
  \end{tabular}
}
\address{$^{1}$Paul C. Lauterbur Research Center for Biomedical Imaging, \\
Shenzhen Institute of Advanced Technology, Chinese Academy of Sciences, Shenzhen, China\\
$^{2}$Peng Cheng Laboratory, Shenzhen, China\\
$^{3}$University of Chinese Academy of Sciences, Beijing, China\\
$^{4}$Department of Computer Science, The University of Hong Kong, Pokfulam, China\\
$^{5}$Pazhou Laboratory (Huangpu), Guangzhou, China
}
\begin{document}
%
\maketitle
\begin{abstract}
Existing contrastive language-image pre-training aims to learn a joint representation by matching abundant image-text pairs. However, the number of image-text pairs in medical datasets is usually orders of magnitude smaller than that in natural datasets. Besides, medical image-text pairs often involve numerous complex fine-grained correspondences. This paper aims to enhance the data efficiency by introducing multiple-to-multiple local relationship modeling to capture denser supervisions. More specifically, we propose a Medical Language-Image Pre-training (MLIP) framework, which exploits the limited image-text medical data more efficiently through patch-sentence matching. Furthermore, we introduce a masked contrastive learning strategy with semantic integrity estimation to reduce redundancy in images while preserving the underlying semantics. Our evaluation results show that MLIP outperforms previous work in zero/few-shot classification and few-shot segmentation tasks by a large margin.
\end{abstract}


\section{Introduction}
\label{sec:intro}

Automated deep learning models for medical image analysis often require large manually labeled datasets during training. Language-image pre-training (LIP)  was proposed to decrease the reliance on manual labels. It leveraged the fact that radiology images are naturally labeled through corresponding clinical reports. These reports can serve as a natural source of weak supervision. With sufficient pre-training, the model can achieve comparable performance with minimal or even no labeling compared with fully-supervised approaches\cite{tiu_expert-level_2022}. However, challenges need to be resolved in medical applications. The first challenge is about the data availability\cite{wang_medclip_2022}. LIP usually requires over millions of image-text pairs with ease\cite{clip_2021}. In the medical domain, only a few hundred thousand pairs are available\cite{johnson_mimic-cxr_2019}. Therefore, addressing the issue of data inefficiency is essential for medical applications. Second, the problem considered in medical diagnosis is naturally fine-grained\cite{huang_gloria_2021}, which requires to distinguish the complex detail relationship between image regions and entities in the report\cite{pan_fine-grained_2023}. Some intuitive ideas involve mining and leveraging additional supervision, either from an external knowledge base or within the image-text pairs themselves such as disease-level semantics\cite{wang_multi-granularity_2022,wang_medclip_2022,liu_improving_2023}, local representations\cite{huang_gloria_2021}, or external knowledge description base\cite{wu_medklip_2023}. But the challenges still remain unresolved due to the complexity of medical data.

\begin{figure}[t]
    \centering
    \includegraphics[width=6cm]{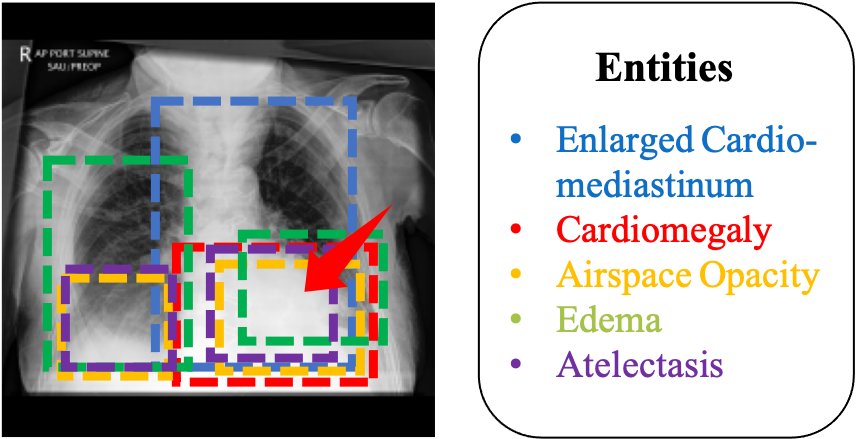}
    \caption{Complex multiple-to-multiple correspondences between medical images and text. An image region can correspond to multiple disease entities, while one disease entity could relate to multiple image regions simultaneously.}
    \label{fig:example}
\end{figure}

This paper aims to address the aforementioned issues with complex multiple-to-multiple local correspondences modeling (Fig. \ref{fig:example}) to catch denser supervision from the image-text pair itself. Thus the data efficiency and fine-grained modeling can be improved. In complex local correspondences modeling, an important factor is the redundant alignment caused by background noise\cite{pan_fine-grained_2023}. We suggest there are two reasons: (1) Images possess substantial information redundancy, making the problem more complex; (2) Text fragment is insufficient to represent semantic information corresponding to the region of interest. We introduce a masked contrastive learning strategy with semantic integrity estimation to reduce the redundancy of image while avoiding the network from learning incorrect correspondences. Moreover, we proposed an sentence-based local alignment approach for compelx correspondences modeling on top of optimal transport theory. Importantly, it does not require any external knowledge base or manual fine-grained labels, thus providing more flexibility. Our results shows that MLIP poses an impressive improvement compare with state-of-the-art methods.

\begin{figure}[t]
    \centering
    \includegraphics[width=8.5cm]{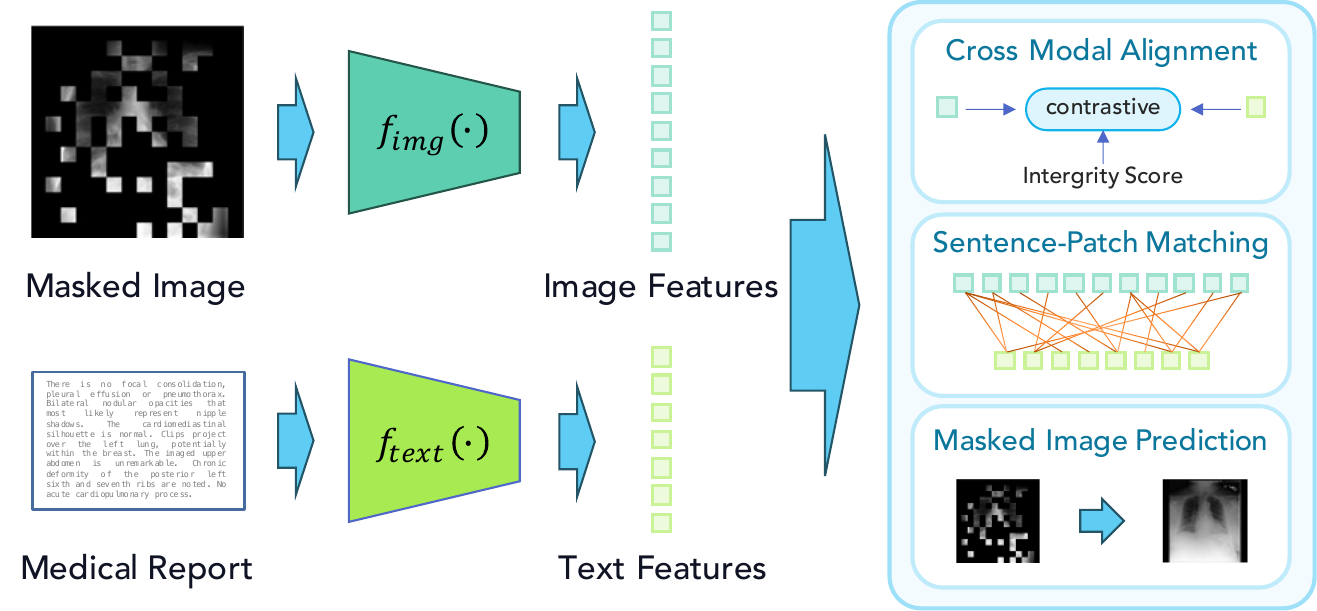}
    \caption{The framework of MLIP.}
    \label{fig:res}
\end{figure}


\section{Methodology}

\subsection{Masked contrastive learning with semantic integrity estimation}
Recent work\cite{pan_fine-grained_2023} has shown that redundant alignments are detrimental to retrieval accuracy. Because these meaningful alignments will be more or less disturbed by other attended fragments irrelevant to the shared semantics. To diminishing the negative effect of redundant alignments, we follow \cite{li_scaling_2023} to incorporate the masking technique into contrastive learning.

MLIP consists of two transformer-based network $f_{img}(\cdot)$ and $f_{text}(\cdot)$to encode masked image patches and text into feature space respectively:
\begin{equation}
{v_i}=f_{img}(\bar{x_i}) ,\quad t_i=f_{text}(y_i)
\end{equation}
Here $y_i \in R^{\left(T+1\right)\times D}$ and $\bar{x_i}=f_{mask}(x_i, M_i) \in R^{\left(L+1\right) \times D}$ are the tokenized text and corresponding masked image patches with an additional [CLS] token.  $L$, $T$, and $D$ represent the number of input image patches, text tokens, and feature dimensions, respectively. We use $f_{mask}(\cdot, M_i)$ to indicate the masking operation where $M_i \in [0,1]^P$ is the mask matrix that was randomly sampled. $P$ is the number of original image patches. In  our implementation, image encoder $f_{img}(\cdot)$ is ViT-B and $f_{text}(\cdot)$ is Bio\_ClinicalBERT.

For global alignment, we use a modified version of contrastive loss to force the network to focus on samples with integral semantics while reduce the impact of false negative samples. The original contrastive loss can be written as:
\begin{equation}
L_{v2t} = -\log\frac{exp(\langle \bar{v_i},\bar{t_i}\rangle/\tau)}{\sum^{N}_{j}exp(\langle \bar{v_i},\bar{t_j}\rangle/\tau)}
\end{equation}
\begin{equation}
L_{t2v} = -\log\frac{exp(\langle \bar{t_i},\bar{v_i}\rangle/\tau)}{\sum^{N}_{j}exp(\langle \bar{t_i},\bar{v_j}\rangle/\tau)}
\end{equation}
where  $\bar{v_i} \in R^D$ is the average of $v_i$ in the first dimension, $\bar{t_i} \in R^D$ is the pooler output of $t_i$. $\tau$ is the temperature factor. There are several benefits of masked contrastive learning. Firstly, masking can effectively reduce the data redundancy of image by removing image patches. Through masking, we can reduce the risk of redundant alignment. Besides, masking in contrastive learning enables us to learn from a larger number of image-text pairs within the same amount of time and contrast more samples per iteration while maintaining a similar memory footprint\cite{li_scaling_2023}.

However, masked images are not always reliable since intrinsic semantics in image might be corrupted. It can lead to unnecessary and even erroneous alignments at both global and local level. We argue that the semantic integrity\cite{bernstein_fast_1989} plays an important role in cross-modal contrastive learning. It ensures that the meaning and relationships between elements in different modalities are effectively captured and aligned during the training. We assume there exists an ideal masking distribution $\Phi \in R^{P}$ which can keep the most of image semantics for most of image in the dataset. In order to achieve this, we employ a learnable matrix $\phi$  to fit such an ideal distribution. Specifically, given a mask matrix $M_i$, we can estimate the integrity score $\hat{w} = \langle\phi, M\rangle \in R^N$ for a batched inputs with batch size $N$, where $\langle\cdot\rangle$ represents the dot-product similarity. The relative semantic integrity score $w_i$ can be computed as:
\begin{equation}
w_i=exp\left(\frac{exp(\hat{w}_i)}{\sum^N_j exp(\hat{w}_j)}\right)
\end{equation}

With relative semantic integrity score $w_i$, we can re-write the cross-modal contrastive loss into:
\begin{equation}
L^*_{v2t} = -\log\frac{exp(w_i*\langle \bar{v_i},\bar{t_i}\rangle/\tau)}{\sum^{N}_{j}exp(w_i*\langle \bar{v_i},\bar{t_j}\rangle/\tau)}
\end{equation}
\begin{equation}
L^*_{t2v} = -\log\frac{exp(w_i*\langle \bar{t_i},\bar{v_i}\rangle/\tau)}{\sum^{N}_{j}exp(w_i*\langle \bar{t_i},\bar{v_j}\rangle/\tau)}
\end{equation}

Compared with the original cross-modal contrastive loss, $L^*_{v2t}$ and $L^*_{t2v}$ can learn more from images with sufficient semantics with an additional temperature scaling, while avoids samples with incomplete semantics.

\subsection{Sentence-patch matching}
As illustrated in Fig.\ref{fig:example}, medical image-text pairs often display significant pairwise complexity with the co-existence of multiple clinical findings. Standard one-to-one contrastive learning is not effective at capturing dense subtle relationships when trained on such complex data. We argue that sufficient semantic information is important in complex local correspondence modeling. Based on this motivation, we introduced a sentence-patch matching (SPM) strategy to modeling local correspondence since each sentence typically describes a single property of the image where token or word may not.  We compute the alignment score between two sets of image patches / text sentences based on optimal transport theory.

Taking a pair of image-text sample features as an example. $v^{'}_i  \in \mathbb{R}^{L \times D}$ and $t^{'}_i \in \mathbb{R}^{T \times D}$ are the patch-level image features and token-level text features. The sentence representation $\hat{t_i} \in \mathbb{R}^{S \times D}$ is computed by aggregating token-level features within $t^{'}_i$ based on sentence segments, where $S$ is the number of sentences in the report.  We use the inexact proximal point method for optimal transports (IPOT) for local alignment quantification. The local align cost matrix $\mathbf{C} \in \mathbb{R}^{L  \times S}$ can be computed by measuring the cosine distance between local features
\begin{equation}
\mathbf{C}_{jk} = 1 - \langle v^{'}_{ij}, \hat{t}_{ik} \rangle_{cos}
\end{equation}

With cost matrix $\mathbf{C}$, the optimal transport strategy matrix $\mathbf{\Gamma} \in \mathbb{R}^{L \times S}$ can be computed with IPOT , i.e. $\mathbf{\Gamma} = \text{iPoT}(\mathbf{C})$. The sentence-patch matching loss can be computed as:
\begin{equation}
L_{SPM}=\mathbf{C} * \mathbf{\Gamma}
\end{equation}

Compared with previous local alignment approach\cite{huang_gloria_2021}, the optimal transport method can calculate local alignment scores without the need for pooling operations. In addition, it does not need fine-grained annotation which alleviates the requirement of manually annotating. We noted that sentence-patch matching loss is a positive-pairs aware loss of dense supervision for cross modal alignment.

\subsection{Masked image prediction}
To further encourages the network to capture fine-grained visual context\cite{huang_gloria_2021} whlie promote holistic understanding beyond low-level image statistics. We add an image decoder on top of image encoder $f_{img}(\cdot)$ to predict masked patches.The masked image prediction loss $L_{MIP}$ was computed by measuring the distance between reconstructed image $x^{'}_i$ and original image $x_i$:
\begin{equation}
L_{MIP} = MSE(x_i,x^{'}_i,M_i)
\end{equation}
where $MSE(\cdot, M_i)$ stands for mean square error function of masked patches. The final loss is:
\begin{equation}
L= \lambda_1 * L^*_{v2t} + \lambda_2 * L^*_{t2v} + \lambda_3 * L_{SPM} + \lambda_4 *L_{MIP}
\end{equation}
where $\lambda_1$, $\lambda_2$, $\lambda_3$ and $\lambda_4$ are weight factors and we set them to 1 for simplicity.


\section{Experiments and Analysis}

\begin{figure}[t]
    \label{fig:vis}
    \begin{minipage}[b]{.48\linewidth}
      \centering
      \label{fig:heatmap}
      \centerline{\includegraphics[width=3cm]{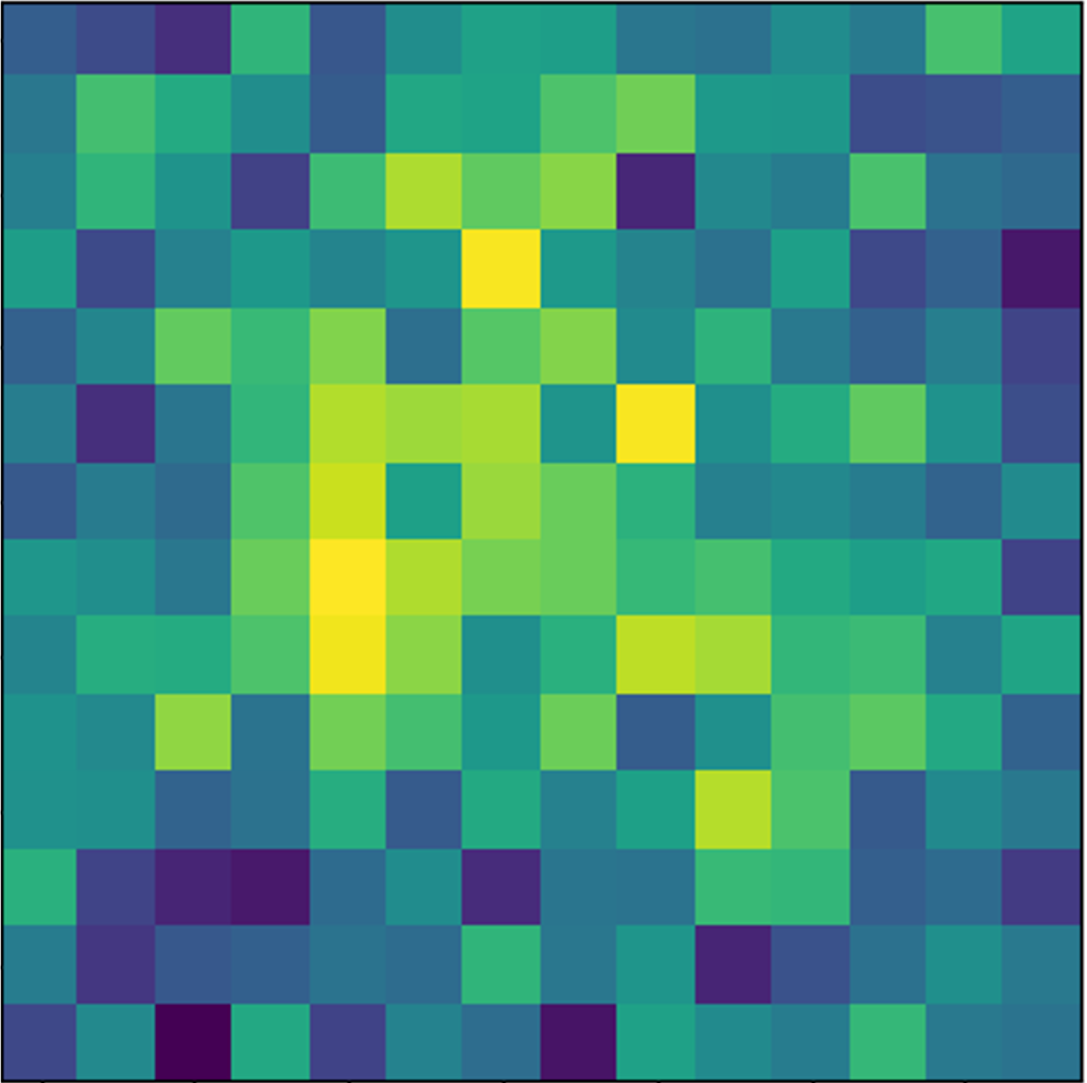}}
      \centerline{(a)}\medskip
    \end{minipage}
    \hfill
    \begin{minipage}[b]{0.48\linewidth}
      \centering
      \label{fig:grounding}
      \centerline{\includegraphics[width=3cm]{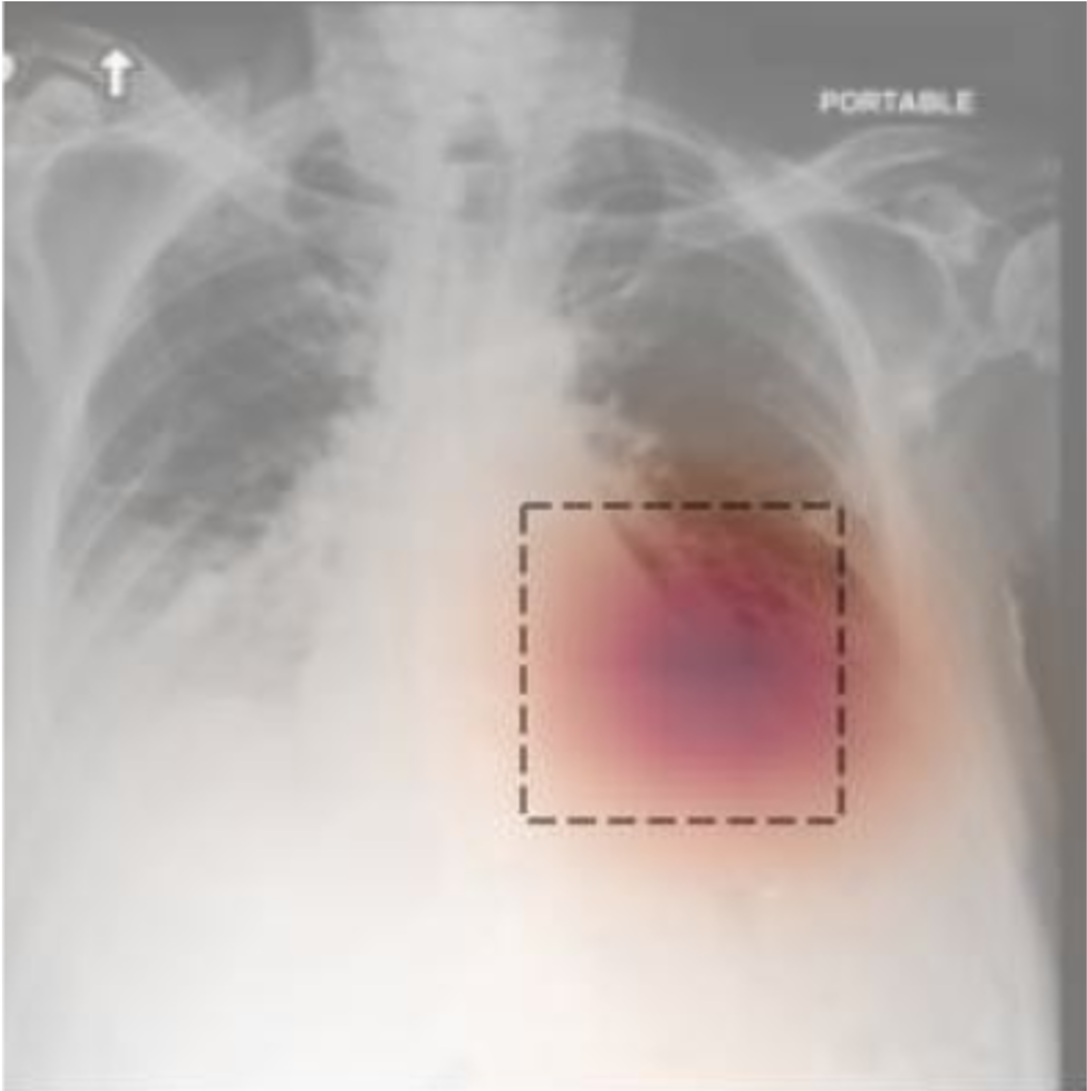}}
      \centerline{(b)}\medskip
    \end{minipage}
    \caption{Visualization of (a) learned $\phi$ and (b) heatmap for a given prompt. The black box in (b) is the corresponding ground-truth. We transform $\phi$ into a square for display.}
\end{figure}

\subsection{Experimental Setup}
We perform pre-training on MIMIC-CXR dataset\cite{johnson_mimic-cxr_2019}. MIMIC-CXR comprises over 370,000 image-text pairs derived from more than 220,000 patient studies. Each image was resized to $224\times224$ with normalization. We divided each image into 196 $14px \times 14px$ patches. For pre-training, we set the masking ratio $r=0.5$ , batch size $N=256$, and temperature $\tau=0.4$. We train our model for 10 epochs with a learning rate 1.5e-4. Linear decay was applied to the learning rate during pre-training by epoch.

\begin{table}[t]
    \centering
    \label{tab:zeroshot}
    \caption{Comparison with state-of-the-art methods on the zero-shot classification task. $^*$ indicates using additional information during pre-training. Results were referenced from \cite{liu_improving_2023} and \cite{wu_medklip_2023}.}
    \setlength{\tabcolsep}{6pt}
    \begin{tabular}{c|cc|cc|c}
        \toprule
        Dataset                              & \multicolumn{2}{c|}{RSNA} & \multicolumn{2}{c|}{SIIM} & {\multirow{2}{*}{AVG}} \\
        Metric                               & AUC              & F1               & AUC                & F1                &                        \\ \midrule
        ConVIRT\cite{zhang_contrastive_2022} & 80.42            & 58.42            & 64.31              & 43.29             & 61.61                  \\
        GLoRIA\cite{huang_gloria_2021}       & 71.45            & 49.01            & 53.42              & 38.23             & 53.03                  \\
        BioVIL\cite{boecking_making_2022}    & 82.80            & 58.33            & 70.79              & 48.55             & 65.12                  \\
        ChexZero\cite{tiu_expert-level_2022} & 85.79            & 63.42            & 68.79              & 47.04             & 66.26                  \\
        SAT\cite{liu_improving_2023}         & -                & 62.00            & -                  & -                 & -                  \\
        MedKLIP$^*$\cite{wu_medklip_2023}    & 86.94            & 63.42            & 89.24              & 68.33             & 76.98                  \\ \midrule
        \textbf{Ours}                                 & \textbf{90.02}   & \textbf{82.51}   & \textbf{89.49}     & \textbf{79.70}    & \textbf{85.43}                  \\ \bottomrule
    \end{tabular}
\end{table}

We validate the generalizability of MLIP on zero-shot classification, few-shot classification, and few-shot segmentation tasks. For zero-shot classification, AUC and F1 scores on RSNA\cite{rsna_dataset} and SIIM\cite{siim-acr-pneumothorax-segmentation} dataset were reported. For few-shot fine-tuning, we fine-tuned the model with a portion of labeled training data. The backbone was frozen during fine-tuning. AUC score on RSNA, NIH Chest X-ray\cite{nih_2017_CVPR}, and SIIM were reported. We follow the setting of \cite{zhou2022advancing} for few-shot segmentation. Dice score on SIIM Pneumothorax dataset were reported. We select 10 state-of-the-arts methods on different tasks including ConVIRT\cite{zhang_contrastive_2022}, GLoRIA\cite{huang_gloria_2021}, BioVIL\cite{boecking_making_2022}, ChexZero\cite{tiu_expert-level_2022}, SAT\cite{liu_improving_2023}, MedCLIP\cite{wang_medclip_2022}, MedKLIP\cite{wu_medklip_2023}, MGCA\cite{wang_multi-granularity_2022}, REFERS\cite{zhou_generalized_2022}, and MRM\cite{zhou2022advancing} for comparision.

\subsection{Classification}

\begin{table}[t]
    \centering
    \label{tab:finetune}
    \caption{Comparison of AUC score with state-of-the-art methods on few-shot fine-tuning classification task. $^\dagger$ indicate that using additional information during pre-training. Results were referenced from \cite{wu_medklip_2023}.}
    \setlength{\tabcolsep}{9pt}
    \begin{tabular}{ccccc}
        \toprule
        Dataset                                 & RSNA           & NIH            & SIIM           & \multirow{2}{*}{AVG} \\
        Data Ratios                             & 1\%            & 1\%            & 10\%           &                      \\ \midrule
        Baseline                                & 71.07          & 60.05          & 61.20          & 64.11                \\
        ConVIRT\cite{zhang_contrastive_2022}    & 83.98          & 66.15          & 78.26          & 76.13                \\
        GLoRIA\cite{huang_gloria_2021}          & 85.99          & 67.10          & 85.38          & 79.49                \\
        BioViL\cite{boecking_making_2022}       & 82.33          & 69.52          & 77.75          & 76.53                \\
        MedKLIP$^\dagger$\cite{wu_medklip_2023} & 87.31          & 77.21          & \textbf{90.71} & 85.08                \\ \midrule
        \textbf{Ours}                                    & \textbf{91.03} & \textbf{78.26} & 90.69          & \textbf{86.66}       \\ \bottomrule
    \end{tabular}
\end{table}

\begin{table}[t]
    \centering
    \label{tab:segmentation}
    \caption{Dice score on SIIM pneumothorax segmentation with 10\% training data. $\dagger$ indicate using additional information during pre-training. $*$ indicate that zero-shot evaluation was not performed in origin paper. Results were referenced from \cite{liu_improving_2023,zhou2022advancing}.}
    \setlength{\tabcolsep}{8mm}
    \begin{tabular}{lc}
    \toprule
    \multicolumn{1}{c}{Method}            & SIIM Pneumothorax \\ \midrule
    \multicolumn{1}{c}{ConVIRT\cite{zhang_contrastive_2022}}     & 43.2                                   \\
    \multicolumn{1}{c}{GLoRIA-CNN\cite{huang_gloria_2021}}  & 46.9                                  \\
    \multicolumn{1}{c}{GLoRIA-ViT\cite{huang_gloria_2021}}  & 71.8                                 \\
    \multicolumn{1}{c}{SAT\cite{liu_improving_2023}}  & 68.2                                 \\ 
    \multicolumn{1}{c}{MedKLIP$^\dagger$\cite{wu_medklip_2023}}     & 72.1                                  \\
    \multicolumn{1}{c}{MGCA$^*$\cite{wang_multi-granularity_2022}}        & 59.3                                 \\
    \multicolumn{1}{c}{REFERS$^*$\cite{zhou_generalized_2022}}      & 72.1                                \\
    \multicolumn{1}{c}{MRM$^*$\cite{zhou2022advancing}}         & 73.2                            \\ \midrule
    \multicolumn{1}{c}{\textbf{Ours}}  & \textbf{73.5}              \\ \bottomrule
    \end{tabular}
\end{table}

Table 1 presents the zero-shot classification result on RSNA and SIIM datasets. MLIP outperforms other methods by over 8.45\% on average score across two datasets. The F1 score of MLIP is significantly higher than the state-of-the-art method MedKLIP by 19.09\% on the RSNA dataset and 11.37\% on the SIIM dataset. That reflects the robustness of MLIP in discriminating positive/negative samples, i.e. image retrieval. In contrast, the local feature alignment approach GLoRIA does not exceed MLIP in zero-shot classification. We refer this to the negative impact of redundant alignment. MLIP can efficiently reduce the impact of redundant alignment and misalignment caused by corrupted semantics.

In addition, we evaluate the performance of the proposed method on few-shot fine-tuning classification task in Table 2. MLIP outperformed the state-of-the-art method on RSNA and NIH Chest X-ray datasets by 3.71\% and 1.05\% respectively. However, MedKLIP was still the best method on the SIIM dataset with a 0.02\% advantage of AUC over MLIP. We noted that additional information was used for MedKLIP during pre-training. 

\subsection{Segmentation}
Table 3 shows the segmentation result on SIIM Pneumontorax dataset. To evaluate the generalizability of learned visual representation, we conduct our experiment with 10\% labeled training data only. MLIP outperforms all of baseline methods on segmentation task. We argue that was benefit from the sentence-patch matching, since it provide dense local-level supervision from original image-text pairs.

\subsection{Ablation Study}
We conduct additional experiments to inspect the internal contribution of each module. We sequentially add each module to the model architecture to assess its impact. The averaged score of zero-shot and few-shot classification was reported in Table 4. Sentence-patch matching takes the most significant improvement by 2.267\% since it can mine dense local supervision from data pairs. Moreover, $L^*_{v2t/t2v}$ with integrity estimation took the second place with 1.76\% improvement, which reveals the importance of semantic complementness of image data in LIP. In addition, we provide the visualization of learned $\phi$ and a sample of the heatmap for a given prompt in Fig. \ref{fig:vis}. MLIP can accurately identify the related region in the image.

\begin{table}[t]
    \label{tab:ablation}
    \caption{Ablation study of MLIP. We report the average score of classification tasks.}
    \centering
    \setlength{\tabcolsep}{9pt}
    \begin{tabular}{cccc|c}
    \toprule
        $L_{v2t/t2v}$ & $L^*_{v2t/t2v}$ & $L_{SPM}$    & $L_{MIP}$    & AVG     \\ \midrule
        $\checkmark$  &     -        &      -       &      -       & 83.814 \\
          -           & $\checkmark$ &      -       &      -       & 85.574 \\
        $\checkmark$  &     -        & $\checkmark$ &      -       & 86.081 \\
          -           & $\checkmark$ & $\checkmark$ &      -       & 86.046 \\
          -           & $\checkmark$ & $\checkmark$ & $\checkmark$ & \textbf{86.557} \\ \bottomrule
    \end{tabular}
\end{table}

\section{Conclusion}
This paper proposes a language-image pre-training framework for the medical application. It is a data-efficient method with local supervision mining that does not need manual labels or external information. Our results demonstrate dense local correspondence is a strong source for the training. In the future, it is expected to adapt the proposed method to more clinical tasks and data modalities.


\clearpage

\section{Compliance with ethical standards}
This research study was conducted retrospectively using human subject data made available in open access. Ethical approval was not required as confirmed by the license attached with the open access data.

\section{Acknowledgement}
This research was partly supported by the National Natural Science Foundation of China (62222118, U22A2040), Shenzhen Science and Technology Program (RCYX20210706092-104034, JCYJ20220531100213029), Guangdong Provincial Key Laboratory of Artificial Intelligence in Medical Image Analysis and Application (2022B1212010011), the major key project of Peng Cheng Laboratory under grant PCL2023AS1-2, and Key Laboratory for Magnetic Resonance and Multimodality Imaging of Guangdong Province (2020B1212060051).

\bibliographystyle{IEEEbib}
\bibliography{strings,refs}

\end{document}